\documentclass[11pt]{article}

\usepackage[final]{acl}
\usepackage{tcolorbox}
\tcbuselibrary{skins}
\usepackage{listings}
\usepackage{times}
\usepackage{latexsym}
\usepackage{xurl}
\usepackage[T1]{fontenc}

\usepackage[utf8]{inputenc}
\usepackage{booktabs}
\usepackage{amssymb}
\usepackage{amsthm}
\usepackage{amsmath}
\usepackage{microtype}
\usepackage{algorithm}
\usepackage{algpseudocode}
\usepackage{xspace}
\newtheorem{definition}{Definition}
\newcommand{\model}{SEARL\xspace}

\usepackage{inconsolata}

\usepackage{listings}
\usepackage{xcolor}

\usepackage{graphicx}
\usepackage{enumitem}
\usepackage{multirow}
\title{SEARL: Joint Optimization of Policy and Tool Graph Memory for Self-Evolving Agents}

\author{\textbf{Xinshun Feng}$^{1 \hypersetup{linkcolor=black}\thanks{Equal contribution.}}$ \quad \textbf{Xinhao Song}$^{1,2 \hypersetup{linkcolor=black}\footnotemark[1]}$ \quad \textbf{Lijun Li}$^{1 \hypersetup{linkcolor=black}\footnotemark[1] \hypersetup{linkcolor=black}\footnotemark[2]}$ \\
  \textbf{Gongshen Liu}$^{2}$ \quad \textbf{Jing Shao}$^{1 \hypersetup{linkcolor=black}\thanks{Corresponding Author}}$ \\
  $^{1}$Shanghai Artificial Intelligence Laboratory \quad $^{2}$Shanghai Jiaotong University\\
  \texttt{\{fengxinshun, songxinhao, lilijun, shaojing\}@pjlab.org.cn} \\
  \texttt{lgshen@sjtu.edu.cn}
}

\begin{document}
\maketitle
\begin{abstract}
Recent advances in Reinforcement Learning with Verifiable Rewards (RLVR) have demonstrated significant potential in single-turn reasoning tasks.
With the paradigm shift toward self-evolving agentic learning, models are increasingly expected to learn from trajectories by synthesizing tools or accumulating explicit experiences.
However, prevailing methods typically rely on large-scale LLMs or multi-agent frameworks, which hinder their deployment in resource-constrained environments.
The inherent sparsity of outcome-based rewards also poses a substantial challenge, as agents typically receive feedback only upon completion of tasks. 
To address these limitations, we introduce a \texttt{Tool-Memory} based self-evolving agentic framework \textbf{\model}. 
Unlike approaches that directly utilize interaction experiences, our method constructs a structured experience memory that integrates planning with execution. 
This provides a novel state abstraction that facilitates generalization across analogous contexts, such as tool reuse.
Consequently, agents extract explicit knowledge from historical data while leveraging inter-trajectory correlations to densify reward signals.
We evaluate our framework on knowledge reasoning and mathematics tasks, demonstrating its effectiveness in achieving more practical and efficient learning\footnote{Code available at \url{https://github.com/circles-post/SEARL}}.
\end{abstract}

\section{Introduction}

\begin{figure}[ht]
    \centering
    \includegraphics[width=\linewidth]{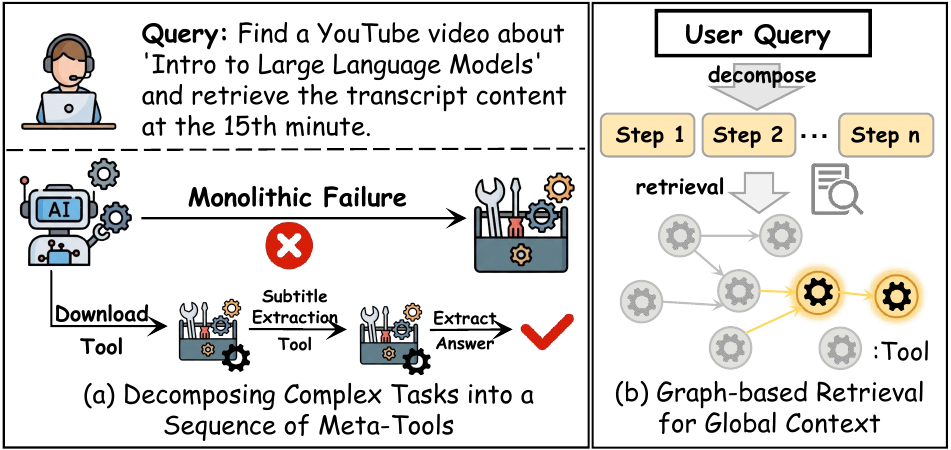}
    \caption{Limitations of existing paradigms. Monolithic tool synthesis overwhelms LLMs, necessitating a divide-and-conquer strategy, while current memory designs suffer from a lack of structural connectivity.} 
    \label{fig:teaser}
\end{figure}
Recent advances in benchmarking agentic capabilities, such as GAIA~\citep{mialon2023gaia}, Humanity's Last Exam (HLE)~\citep{phan2025humanity}, and WebArena~\citep{zhou2023webarena}, have revealed that direct prompting for large language models (LLMs) remains insufficient 
for solving complex, long-horizon tasks~\cite{hu2025owl, lu2025octotools}. 
A central challenge in this paradigm is determining the tool set available to the agent: the availability of appropriate tools is often a prerequisite for task completion.
Most existing frameworks~\cite{yao2023react} adopt a static design, predefining a large, fixed tool list from which the agent must select and invoke tools, potentially limiting adaptability and generalization across diverse tasks.

Recent works typically address this challenge from two perspectives.
First, tool-generation methods like Alita~\cite{qiu2025alita} and STELLA~\cite{jin2025stella} autonomously create tools but store them in unstructured repositories.
This lack of structure leads to high-level abstractions, limiting both reusability and fine-grained composition.
Furthermore, given the limited reasoning capabilities of smaller-scale LLMs, generating monolithic tools often results in failure; consequently, decomposing complex tasks into subtasks and creating corresponding tools proves to be a more effective strategy.
Second, RL-based or experience-driven approaches~\cite{tang2025agent} leverage past trial data but often overlook the explicit dependency relations essential for complex reasoning. 
To bridge these gaps, we propose the \texttt{Tool Graph}, a structured memory where tools serve as nodes and execution dependencies as edges (Figure \ref{fig:teaser}).
This graph evolves continuously, providing strong inductive biases to improve generalization and planning.

While agentic reinforcement learning~\cite{singh2025agentic} has emerged as a prominent paradigm, it faces significant limitations.
First, their reward designs typically prioritize trajectory-level success or format correctness, largely neglecting step-level feedback on reasoning quality. 
Although some methods incorporate process-level rewards, they often rely on heuristic designs or are tailored to specific domains~\cite{feng2025group}, which restricts their applicability in general scenarios.
Second, they focus on improving the intricacies of the models, ignoring the potential of expanding external memory to achieve long-term improvement. 
Motivated by these gaps, we propose \model, a reinforcement learning framework in which both the policy model and the tool-based memory evolve jointly during training.

Our framework enables the joint evolution of the agent's memory and policy, surpassing methods that optimize either component in isolation. 
It comprises two key components: (i) An environment augmented with a dynamic \texttt{Tool Graph} that supports continuous tool creation and reuse, serving as an evolving structured memory that progressively accumulates problem-solving capabilities.
(ii) A tool-memory-aware policy optimization algorithm fine-tuned via trajectory- and step-level credits, which adapts the LLM to effectively navigate and leverage the growing graph structure.
Through this joint optimization, the agent exhibits \emph{self-evolution}, becoming increasingly competent as it encounters and solves more complex tasks.
In summary, the key contributions of this work are as follows:
\begin{itemize}
    \item We introduce \textbf{\model}, a new paradigm that jointly optimizes both policy parameters and external tool-based memory, enabling agents to continuously acquire, refine, and reuse problem-solving capabilities.
    \item We propose a \texttt{Tool-Memory-Aware} training algorithm that extends step-level advantages with memory-anchored clustering, providing fine-grained credit assignment for tool creation and execution.
    \item We formalize the \texttt{Tool Graph Memory} as a structured, persistent representation of tool knowledge and inter-tool dependencies, and show that its growth improves generalization and planning in complex tasks.
\end{itemize}
Experimental results validate the effectiveness of our method, showing that it empowers small LLMs with robust self-evolving capabilities driven by an ever-expanding tool memory.

\section{Preliminary}

\begin{definition} [Self-Evolving Agent]
A self-evolving agent improves its problem-solving capabilities through experience. 
This evolution typically follows two paradigms: \textbf{tool-based evolution}, which optimizes a set of tools $\mathcal{T}$ for specific tasks, and \textbf{memory-based evolution}, which leverages a repository of past successful trajectories $\mathcal{M}$.
Ideally, this continuous adaptation enables the agent to generalize its reasoning processes to novel, unseen states without requiring manual retraining.
\end{definition}

\begin{definition} [Tool Memory-Enhanced MDP]
We define a Tool Memory-Enhanced Markov Decision Process as a tuple $\langle \mathcal{S}, \mathcal{A}, \mathcal{P}, \mathcal{R}, \mathcal{T}_{G} \rangle$. Here, $\mathcal{S}$ denotes the state space, $\mathcal{A}$ the action space, $\mathcal{P}:\mathcal{S}\times\mathcal{A}\rightarrow \Delta(\mathcal{S})$ the state transition dynamics, and $\mathcal{R}:\mathcal{S}\times\mathcal{A}\rightarrow\mathbb{R}$ the reward function. 
Crucially, $\mathcal{T}_{G}=(V, E)$ represents the graph-structured tool memory, where $V \subset \mathcal{S}\times\mathcal{A}\times\mathbb{R}$ denotes the set of stored interaction tuples, and $E$ encodes correlations derived from trajectories.
\end{definition}

\paragraph{Problem Formulation.}
Given a task description $x$ drawn from distribution $\mathcal{D}$ at time step $t$, the agent observes state $s_t \in \mathcal{S}$ and generates a textual action $a_t$, and transitions to $s_{t+1}$.
A complete interaction trajectory is denoted as $\tau = \{(s_i,a_i,r_i)\}^N_i$.
The agent operates under an LLM-based policy $\pi_\theta(a_t|s_t,x,\mathcal{T}_{G})$, parameterized by $\theta$. 
The training objective seeks to maximize the expected reward while regularizing deviation from a reference policy $\pi_{\text{ref}}$ via KL divergence:

\begin{equation}
\begin{split}
& \max_{\pi_\theta} \mathbb{E}_{x \sim \mathcal{D}, y \sim \pi_\theta(\cdot \mid x; \mathcal{T}_{G})}
\left[ r_\phi(x, y) \right]
- \\
& \beta \, \mathbb{D}_{\text{KL}} \left[ \pi_\theta(y \mid x; \mathcal{T}_{G}) \,\|\, \pi_{\text{ref}}(y \mid x; \mathcal{T}_{G}) \right],
\end{split}
\end{equation}

The trajectory generation process $p(\tau)$ is decomposed into four sequential phases:
(1) \textbf{Retrieve}: A retrieval function $\mu$ selects the most relevant tool $T_t$ from memory $\mathcal{T}_G$;
(2) \textbf{Reuse}: The agent decides whether to utilize the retrieved tool for the current action;
(3) \textbf{Creation}: The agent determines if a new, universal tool should be defined and added to memory;
and (4) \textbf{Transition}: The environment evolves to the next state based on the dynamics.
Formally, this decomposition is expressed as:
\begin{equation}
\begin{split}
    &p(\tau) = \prod_{t=0}^{N-1} 
    \underbrace{ \mu(T_t | s_t,\mathcal{T}_G) }_{(1)\text{Retrieve}} \, 
    \underbrace{ p_{LLM}(a_t | s_t,T_t) }_{(2)\text{Reuse}} \\
    &\times 
    \underbrace{ p_{LLM}(T_{t+1} | a_t ,T_t) }_{(3)\text{Creation}} \,
    \underbrace{ \mathcal{P}(s_{t+1} | s_t,a_t) }_{(4)\text{Transition}}
\end{split}
\end{equation}
where $N$ denotes the maximum steps, and $T_t$ is the pertinent tool retrieved from $\mathcal{T}_G$.

\begin{figure*}[!ht]
  \centering
  \includegraphics[width=\textwidth]{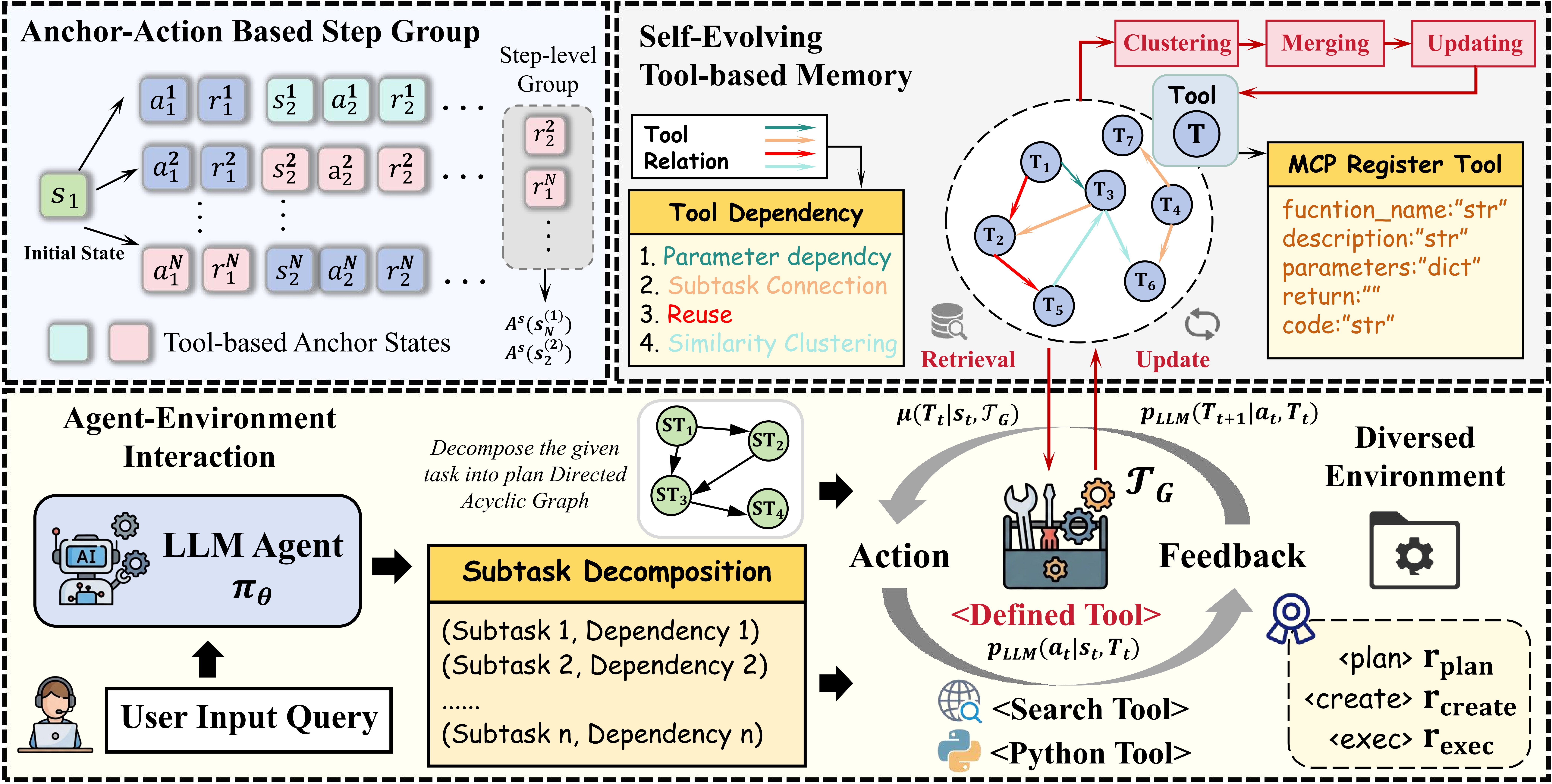}
  \caption{Overview of the proposed framework. The architecture consists of three components: (1) Agent-Environment Interaction (Bottom): decomposes tasks and generates structured trajectories via planning, retrieval, and execution. (2) Anchor-Action Step Grouping (Top-Left): leverages tool-based anchors to enable fine-grained, step-level advantage estimation. (3) Self-Evolving Tool-based Memory (Top-Right): A graph-structured memory $\mathcal{T}_G$ that dynamically retrieves, clusters, and updates MCP tools based on trajectory dependencies.}
  \label{fig:my-figure-label}
\end{figure*}
\section{Self-Evolving Agent Reinforcement Learning}
This section details the proposed method, including the overall workflow, the design of the reward function, the advantage estimation and policy optimization algorithm, and the tool-based memory structure for retrieving and reusing relevant tools.\par

\subsection{Structured Trajectory Generation}   
Our approach formalizes the decision-making process as a structured sequence of meta-reasoning stages, explicitly annotated to enable fine-grained control.
Initially, a global plan is generated to delineate the high-level strategy and the execution order of subtasks.
Departing from methods that treat subtask execution as a monolithic opaque block, we decompose the process into distinct phases encompassing tool retrieval, reasoning, and external actions.
Specifically, we define four step-level components, each delimited by XML-style tags (e.g., \texttt{<tool\_call>}), with the entire sequence encapsulated within a \texttt{<subtask>} tag.
\begin{itemize}
    \item \textbf{Planning}: Decomposes the task into high-level steps to formulate an overall strategy at the start of a task.
    \item \textbf{Retrieve}: Selects the most relevant tool $T_t$ from the tool-based memory $\mathcal{T}_G$ using a retrieval policy $\mu(T_t \mid s_t,\mathcal{T}_G)$, 
providing context for the upcoming action.
    \item \textbf{Think}: Performs internal deliberation conditioned on the state, deciding whether to use a specific tool or produce a direct answer.
    \item \textbf{Action}: Generates an output, which can be either an \texttt{<answer>} (a textual response) or a \texttt{<tool\_call>} command for code execution. 
    To simplify tool reuse and unify the creation process, the MCP tool creation is also defined as a regular tool calling.
\end{itemize}


\subsection{Reward Shaping}   
During reinforcement learning, the agent is guided by a composite reward signal that combines task completion with feedback on the agent's interaction with tool-based memory. 
The overall signal consists of a sparse outcome reward that reflects whether the final answer solves the task, and a dense, process-level reward tied to planning, tool usage, and code execution. 

\paragraph{Outcome Reward ($R(\tau)$).} A binary signal awarded at the conclusion of a trajectory. 
Specifically, $R(\tau)=r_s$ if the task is successfully completed with the correct answer, and $0$ otherwise.
In our experimental setting, the constant $r_s$ is set to $1$.

\paragraph{Behavioral Reward ($r_t^{\tau}$)} We use a dense reward, assigned at each step t, to incentivize locally beneficial behaviors. 
To guide the agent's behavior toward a desired direction, we design distinct rewards for different actions.
\begin{itemize}
    \item \textbf{Planning Reward ($r_\text{planning}$)}: Awarded if the generated plan can be successfully parsed into a complete sequence of subtasks with executable subplans.
    \item \textbf{Tool Creation Reward ($r_\text{creation}$)}: Granted when the agent creates a new MCP tool that conforms to the required registration format, encouraging a meaningful extension of tool-based memory $\mathcal{T}_G$.
    \item \textbf{Tool Execution Reward ($r_\text{execution}$)}: The tool call generated is awarded when it executes successfully and returns a valid output, promoting reliable and verifiable behavior.
\end{itemize}

\paragraph{Format Reward ($r_t^\text{format}$).} 
A positive reward $+\lambda_{\text{format}}$ is granted at step $t$ if the model output conforms to the required structure.

\subsection{Advantage Estimation with Tool-Memory-Aware Policy Optimization}    
While group-based RL has proven effective for single-turn tasks, extending it to multi-step agent settings faces significant credit assignment challenges. 
Existing step-level approaches~\cite{feng2025group} attempt to mitigate this by grouping identical states. 
However, this relies heavily on frequent state re-visitation (e.g., in GUI environments). 
In more general, open-ended environments, the state space is vast and continuous, rendering such precise state matching impractical. 
Compounding this issue, basic step-level reward designs are susceptible to reward hacking, often lacking a reliable correlation with actual reasoning quality. 
Furthermore, current frameworks offer limited mechanisms for auditing these risks or implementing effective mitigation measures.

To overcome these limitations, we propose our algorithm in this section, combining step-level rewards with tool-based memory $\mathcal{T}_G$. 
Instead of grouping by raw environment states, we define the tool utilization as the \textbf{anchors} during advantage computation.
We leverage a two-level advantage structure: (i) episode-level relative advantages capture the global effectiveness of entire trajectories, providing a stable task-level learning signal, and (ii) tool-based memory-anchored step advantages deliver fine-grained credit for tool creation, reuse, and tool-execution decisions.

\paragraph{Episode-Level Relative Advantages.}
The episode-level relative advantage is computed over a group of $N$ trajectories $\{\tau_i\}_{i=1}^{N}$ rolled out under the same task and initial state.
For each trajectory, we utilize the total return $R(\tau_i) = R_{\mathrm{orm}}(\tau_i) + \sum_{t=1}^{T} r^{(i)}_t$ as a holistic measure of task completion quality, where $R_{\mathrm{orm}}$ denotes the rule-based score and $\sum_{t=1}^{T} r^{(i)}_t$ aggregates process-level rewards.
The resulting set of trajectory-return pairs forms an \emph{episode-level} group:
\begin{equation}
    \mathcal{G}^E = \left\{ (\tau_i, R(\tau_i)) \right\}_{i=1}^{N}
\end{equation}
The episode relative advantage $A^E(\tau_i)$ for each $\tau_i$ can be formalized as:
\begin{equation}
    A_E(\tau_i) = 
    \frac{R(\tau_i) - \frac{1}{N}\sum_{j} R(\tau_j)}
    {\sqrt{\frac{1}{N}\sum_{j}\left(R(\tau_j) - \frac{1}{N}\sum_{k} R(\tau_k)\right)^2}}
\end{equation}

\paragraph{Step-Level Relative Advantages.} 
While the episode-level relative advantage provides a coarse signal, it cannot distinguish the contributions of individual actions within a trajectory.
To provide fine-grained credit assignment, we construct step-level groups using tool-based memory anchors rather than raw environment states. 
Specifically, let $\mathcal{U} = \{g_1, g_2, \dots, g_{|\mathcal{U}|}\}$ denote the set of all distinct MCP tools appearing across the trajectory group $\{\tau_1, \tau_1, \dots,\tau_N\}$.
For each MCP tool $g$ in $\mathcal{U}$, we collect all actions associated with $t$ into a group. 
\begin{equation}
    \mathcal{G}_S(g) = \left\{ \left(a^{(i)}_t, R^{(i)}_t\right) \right\}_{(i,t) \in \mathcal{I}_g}
\end{equation}
where $R^{(i)}_t = \sum_{k=t}^{T} \gamma^{k-t} r^{(i)}_k$ denotes the discounted return-to-go for the $i$-th trajectory starting from step $t$ drawn inspiration from~\cite{feng2025group}.
After all trajectories are generated, we perform a post-processing merge operation on the tool-based memory $\mathcal{T}_G$: newly created MCP tools are compared against existing ones using a similarity metric over their name and description. 
If the similarity exceeds a threshold, the new tool is merged with the most similar existing tool, and $\mathcal{T}_G$ is updated accordingly.
This procedure defines a unique memory anchor $g$ for each equivalence class of tools.
All steps interacting with anchor $g$ are aggregated into the same step-level group $\mathcal{G}_S(g)$. 
This grouping unifies credit assignment across trajectories and temporal contexts, ensuring that advantage estimation captures the specific utility of the MCP tool, isolated from unrelated contextual variances.
Once these step-level groups are formed, the \emph{step relative advantage} for each $g\ \sim\ \mathcal{U}$ and each action $a_t^{(i)} \in G^S(g)$ can be formalized as:
\begin{equation}
\begin{aligned}
    \mu_{\mathcal{G}} &= \frac{1}{|\mathcal{G}_S(g)|}\sum_{(a,r)\in \mathcal{G}_S(g)} r, \\
    A_S(a^{(i)}_t) &= \frac{R^{(i)}_t - \mu_{\mathcal{G}}}
    {\sqrt{\frac{1}{|\mathcal{G}_S(g)|}\sum_{(a,r)\in \mathcal{G}_S(g)}\left(r - \mu_{\mathcal{G}}\right)^2}}
\end{aligned}
\end{equation}
where the denominator provides standard-deviation normalization over the group returns.

This tool-based, memory-anchored advantage evaluates the relative utility of actions associated with the same MCP tool across varying trajectories. 
While agents address distinct tasks and generate diverse trajectories, they operate within analogous state subspaces when utilizing the same specific tool.
Consequently, by integrating MCP tool usage into training, we effectively abstract the unbounded real-world state space into a finite set defined by the toolset $\mathcal{T}_G$.
When combined with the episode-level advantage, this approach offers a complementary optimization signal: the episode-level term provides a coarse, trajectory-wide guide, whereas the tool-level advantage assigns fine-grained credit specifically to tool-related decisions. Detailed policy optimization can be found in Appendix~\ref{appendix:policy_opt}.


\subsection{Tool-Based Memory}
In this section, we formally describe the operation and evolution of our \texttt{Tool-Based Memory}, as a directed graph $\mathcal{T}_G = (V, E)$, where nodes $V$ denote registered MCP tools and edges $E$ encode step-level dependencies. 
Serving as an external memory, the lifecycle of this graph encompasses four phases: \textbf{Subgraph Extraction}, \textbf{Tool Registration}, \textbf{Tool Retrieval}, and \textbf{Memory Update}.

\paragraph{Subgraph Extraction.}
During the \texttt{plan} phase, the task is decomposed into subtasks $\{ \mathit{ST}_k \}_{k=1}^m$, forming a dependency graph $G_{\mathrm{plan}}$. 
We project this structure onto the tool space via a mapping $\phi$ to derive a task-specific memory subgraph $\mathcal{T}_G^{(i)}$:
\begin{equation}
\begin{split}
    \mathcal{T}_G^{(i)} &= \bigl(V^{(i)}, E^{(i)}\bigr), \\
    V^{(i)} &= \bigl\{\phi(ST_k) \mid k = 1, \dots, m \bigr\}, \\
    E^{(i)} &= \bigl\{(\phi(u), \phi(v)) \mid (u, v) \in E_{\mathrm{plan}} \bigr\}
\end{split}
\end{equation}
where $E^{(i)}$ preserves the trajectory-level execution order. 
Crucially, we instruct the model to generate dedicated, modular tools for specific subtasks rather than monolithic solvers. 
This granularity not only improves training stability but also ensures the resulting tools capture foundational operations, enhancing their reusability across diverse tasks.

\paragraph{Tool Registration and Retrieval.}
Tool registration is facilitated by the \texttt{mcp creation} tool.
During trajectory execution, whenever the agent invokes this tool, we verify its execution status; successfully executed instances are tentatively added to a candidate pool.
At the end of each training iteration, to prevent redundancy, we calculate cumulative rewards across rollouts and select only the tools associated with the highest rewards for final registration.
For retrieval, given a sequence of decomposed subplans $[p_1, p_2, \dots, p_n]$, we employ a dedicated model to identify the most relevant tools by evaluating the alignment between the content of each subplan and the tool descriptions.
Detailed procedures are provided in Appendix~\ref{appendix:training_Retrieval}.

\paragraph{Memory Update and Consolidation.}
Once task-specific subgraphs $\{\mathcal{T}_G^{(i)}\}$ are constructed, they are integrated into the global memory $\mathcal{T}_G$ through a unified merge operation.
This process involves two parallel mechanisms: semantic node merging and structural edge consolidation.
First, to determine tool equivalence, we compute the semantic embedding $\mathbf{e}(v)$ for each tool.
The similarity between a new tool $v \in V^{(i)}$ and an existing tool $v' \in V$ is measured via the cosine similarity of their normalized embeddings: $\mathrm{sim}(v,v') = \tilde{\mathbf{e}}(v)^\top \tilde{\mathbf{e}}(v') \in [-1,1].$
If $\mathrm{sim}(v,v') \ge \delta$, $v$ is merged with $v'$; otherwise, $v$ is registered as a new node, $\delta$ is a predefined threshold.
Simultaneously, directed edges representing trajectory-level precedence (i.e., subtask $ST_p$ precedes $ST_q$) are incorporated to preserve causal structures.
Crucially, when tools are merged, their incident edges are automatically redirected to the consolidated node, effectively accumulating dependency patterns across diverse trajectories.
The global memory update is formalized as:
\begin{equation}
    \mathcal{T}_G \leftarrow 
    \mathrm{Merge}\!\left(\mathcal{T}_G, 
    \{\mathcal{T}_G^{(i)}\}_{i=1}^{N}\right),
\end{equation}
This procedure ensures that $\mathcal{T}_G$ evolves into a persistent repository that captures both tool-level semantics and sequential dependencies.
The detailed algorithm is provided in Appendix~\ref{appendix:training_Algorithms}.


\section{Experiment}
\subsection{Datasets}
To comprehensively evaluate the effectiveness of our \model  \ in training agents using a tool, we conduct experiments on the following three types of long-horizon reasoning tasks:
\begin{itemize}
    \item \textbf{Mathematical~Reasoning}:  AIME2024 \cite{aime2024}, MATh500~\cite{lightman2023let}, and GSM8K~\cite{hendrycks2021measuring}.
    \item \textbf{Knowledge-Intensive Reasoning}: including WebWalker~\cite{wu2025webwalker}; as well as three Wikipedia-based open-domain QA tasks: HotpotQA~\cite{yang2018hotpotqa}, 2WikiMultihopQA~\cite{ho2020constructing}, and Musique~\cite{trivedi2022musique}, and bamboogle~\cite{press2022measuring}.
\end{itemize}
Following ARPO~\cite{dong2025agentic}, we adopt the same data split settings for all benchmarks, ensuring consistency and comparability of results.

\subsection{Baselines}
To evaluate the effectiveness, we compare \model with common trajectory-level RL algorithms for training LLM-based tool-use agents, including TIR Prompting, GRPO~\cite{shao2024deepseekmath}, DAPO~\cite{yu2025dapo}, REINFORCE++~\cite{hu2025reinforce++}, and ARPO~\cite{dong2025agentic}.
GiGPO~\cite{feng2025group} is excluded from the experiments due to its incompatibility with our task settings.

\subsection{Training Settings}
For mathematical and multi-hop knowledge reasoning tasks, we utilize the 10,000 open-source RL training samples from Tool-star~\cite{dong2025tool} as our training dataset.
The agent is equipped with two fundamental tools: a Python interpreter and a Wikipedia search interface, which is modified based on~\citet{chai2025rlfactoryplugandplayreinforcementlearning}.
Notably, to ensure resource efficiency, we implement the local Wikipedia search server proposed in~\cite{jin2025search}.
Detailed training configurations and specific prompts are provided in the Appendix~\ref{appendix:training_details}.

\subsection{Evaluation Metrics}
For all benchmarks, we adopt an LLM-as-Judge evaluation to ensure a consistent and reliable assessment across diverse tasks.
Specifically, we employ Qwen3-32B as the judge model to assess binary correctness against ground-truth solutions, for its high alignment with human evaluation.
The corresponding prompt is provided in Appendix~\ref{appendix:prompt}.
We report results using pass@1 accuracy.
Model predictions are post-processed by isolating the content between \texttt{<answer>} and \texttt{</answer>} tags, followed by extracting from \verb|\boxed{...}|.
\begin{table*}[th]

\centering
\resizebox{\textwidth}{!}{
\begin{tabular}{l|ccc|cccc|c}
\hline
\multirow{2}{*}{Methods} & \multicolumn{3}{c|}{Mathematical Reasoning} & \multicolumn{4}{c|}{Multi-Hop QA} & \multirow{2}{*}{Avg Rank} \\ \cline{2-8}
 & GSM8K & MATH500 & AIME24 & HotpotQA & 2wiki & Musique & Bamboogle & \\ \hline
TIR Prompt   & 0.2259 & 0.0540 & 0.0000 & \underline{0.2300} & 0.1250 & 0.0350 & 0.1200 & 5.29 \\
GRPO         & \textbf{0.8870} & \textbf{0.7360} & \underline{0.1333} & 0.2150 & 0.3450 & \textbf{0.0900} & 0.1600 & \underline{2.43} \\
DAPO         & 0.8059 & 0.5520 & \underline{0.1333} & \textbf{0.3350} & \underline{0.3500} & \underline{0.0650} & \underline{0.2480} & 3.00 \\
Reinforce++  & \underline{0.8658} & 0.6800 & 0.1000 & 0.1100 & 0.2600 & 0.0000 & 0.0080 & 4.57 \\
ARPO         & 0.8241      & 0.6480      & \textbf{0.3333}      & 0.1400      & 0.2200      & \underline{0.0650}      & 0.1760      & 3.57 \\
\model        & 0.8620      & \underline{0.6820}      & \textbf{0.3333}      & \textbf{0.3350}      & \textbf{0.3600}      & \textbf{0.0900}      & \textbf{0.3040}      & \textbf{1.43} \\ \hline
\end{tabular}
}
\caption{Performance comparison across mathematical reasoning and multi-hop QA benchmarks. \textbf{Bold} indicates the best performance, while \underline{underlined} denotes the second best. The Avg Rank column represents the average ranking across all tasks (lower is better).}
\label{tab:inference-results}
\end{table*}
\subsection{Main Results}

Table~\ref{tab:inference-results} presents the comparative performance of \model\ against strong baselines across mathematical reasoning and multi-hop question answering benchmarks. 
The results indicate that our self-evolving tool-based framework outperforms policy gradient methods in knowledge-intensive reasoning tasks, while demonstrating superior generalization capabilities in complex mathematical problem-solving, thereby distinguishing itself.

\paragraph{Superiority of Structured Memory in Multi-Hop Reasoning.} 
On multi-hop QA datasets (HotpotQA, 2wiki, Bamboogle), \model\ consistently outperforms or matches strong baselines. 
This gap is pronounced in tasks requiring the composition of information from disjoint sources.
We attribute this to the Tool Graph Memory, acting as a persistent external knowledge structure.
A challenge in our setting is that the local Wikipedia search retrieves massive irrelevant context.
Baselines (e.g., GRPO), relying on transient context windows, struggle to filter this noise, often updating policies based on noisy trajectories.
In contrast, our agent leverages graph structure to decompose queries and isolate precise evidence.
This structured retrieval ensures intermediate steps are grounded, significantly reducing hallucinations and maintaining coherence.

\paragraph{Robustness and Generalization in Mathematical Reasoning.} 
In the mathematical domain, results highlight a trade-off: robustness on standard tasks versus exceptional generalization on complex problems.
On benchmarks like GSM8K and MATH500, \model\ remains competitive.
We acknowledge that for simpler problems, autonomous tool generation may introduce minor procedural noise where monolithic reasoning suffices.
However, this slight overhead is justified by the model achieving the highest performance on AIME24, matching that of ARPO.
On this benchmark requiring novel solution paths, our method establishes a significant lead.
While baselines may overfit to static patterns of easier datasets, our self-evolving mechanism dynamically constructs tools to decompose intricate problems, demonstrating superior adaptability.

\subsection{Ablation Study and Analysis}
In this section, we investigate the impact of the training algorithm on learning dynamics and evaluate the contribution of each component.

\begin{figure}[h]
  \centering
  \includegraphics[width=\linewidth]{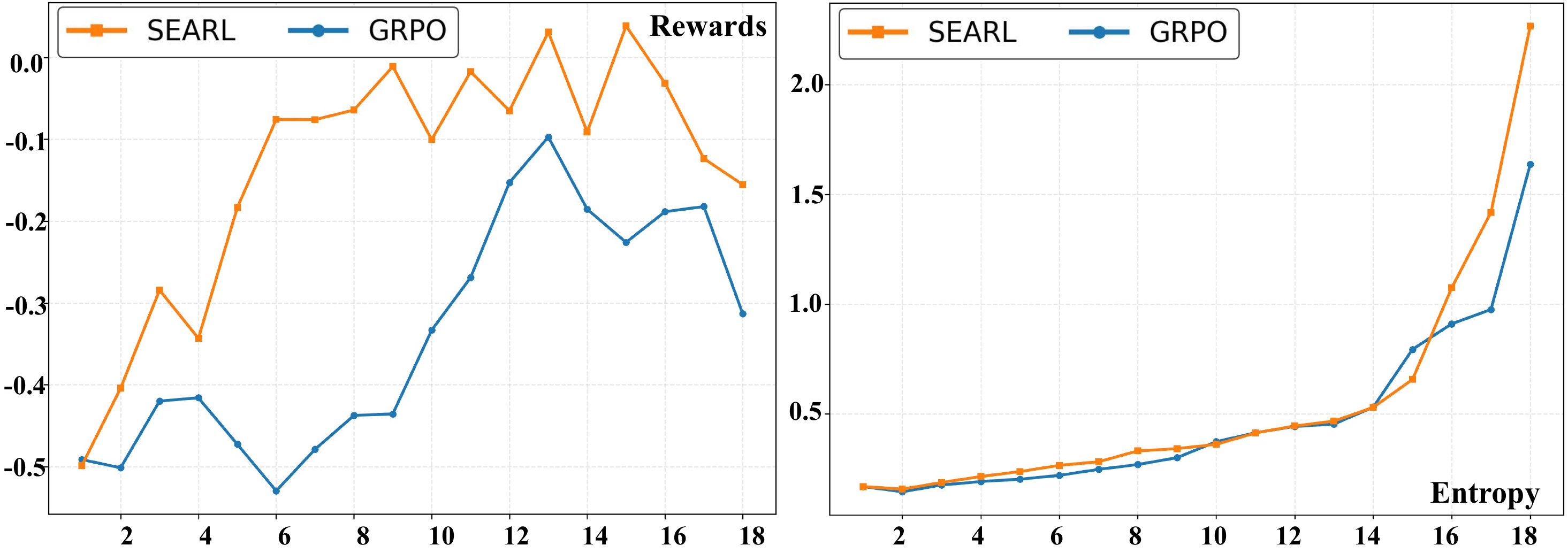}
  \caption{Comparison of training rewards and entropy between GRPO and \model.}
  \label{fig:dynamic}
\end{figure}

\textbf{Learning Dynamics.} 
Figure~\ref{fig:dynamic} illustrates the evolution of overall training rewards and entropy, benchmarking \model against the GRPO baseline under identical workflow conditions.
\model training rewards outperforms GRPO throughout the process, suggesting that it effectively leverages step-grouped advantages to derive more informative feedback. 
\model maintains higher entropy levels throughout training, indicating sustained exploration capabilities. 
Notably, the training rewards remain predominantly negative.
Drawing inspiration from~\cite{lee2025banel}, we deliberately impose strict negative penalties to deter redundant tool invocations and failed creation attempts.

\begin{figure}[h]
  \centering
  \includegraphics[width=1\linewidth]{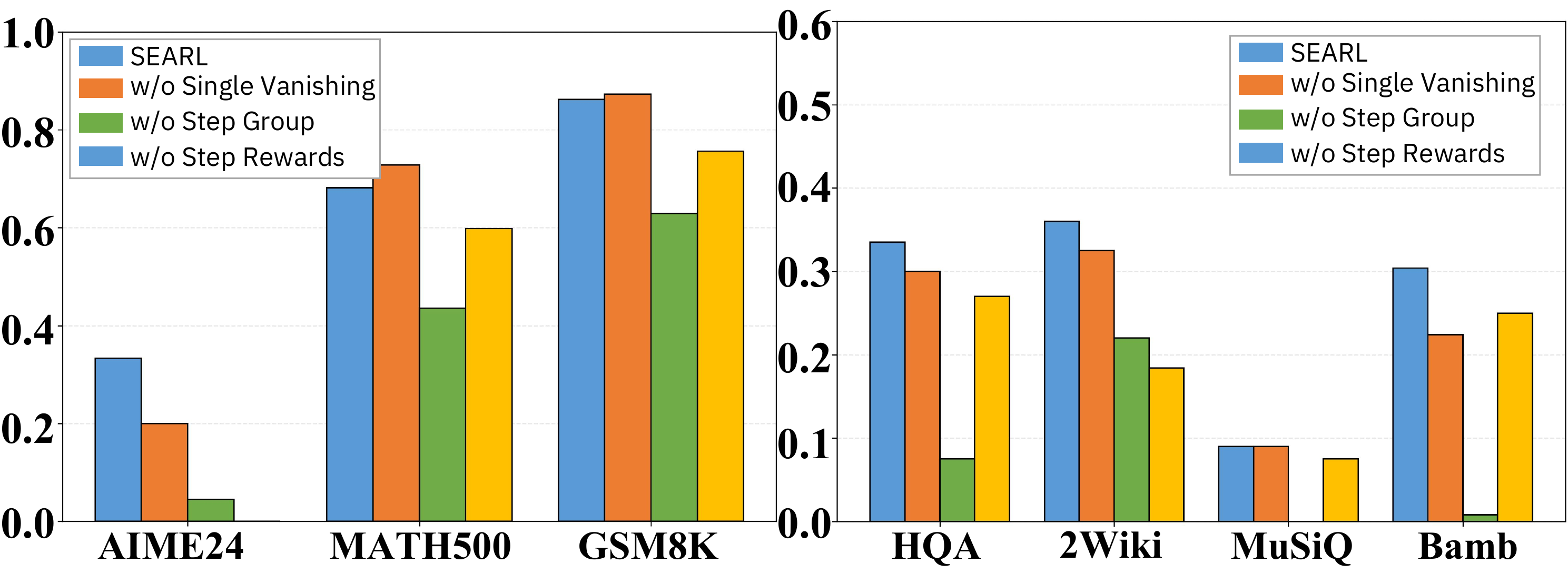}
  \caption{Ablation study illustrating the impact of removing different components.}
  \label{fig:ablation_study}
\end{figure}

\textbf{Component Analysis.} 
To validate the effectiveness of our design, we conduct an ablation study by selectively removing three key components: (i) \textit{Single Vanishing}, which bypasses group-level advantage estimation when a group contains only a single element; (ii) \textit{Step-level Grouping}, which removes the grouping strategy entirely; and (iii) \textit{Step Rewards}, which eliminates fine-grained process-level reward feedback.
The results are presented in Figure~\ref{fig:ablation_study}.
We observe that removing Step-level Grouping leads to the most significant performance degradation across the majority of datasets (e.g., AIME24 and Bamb), underscoring its critical role in accurate advantage estimation. 
Similarly, the absence of Step Rewards results in a noticeable drop, confirming the necessity of dense supervision signals. 
In contrast, while the Single Vanishing mechanism has a relatively smaller impact, it remains essential for maintaining overall stability.

\subsection{Implementation of the Tool Graph}
\begin{figure*}[!ht]
  \centering
  \includegraphics[width=0.9\textwidth]{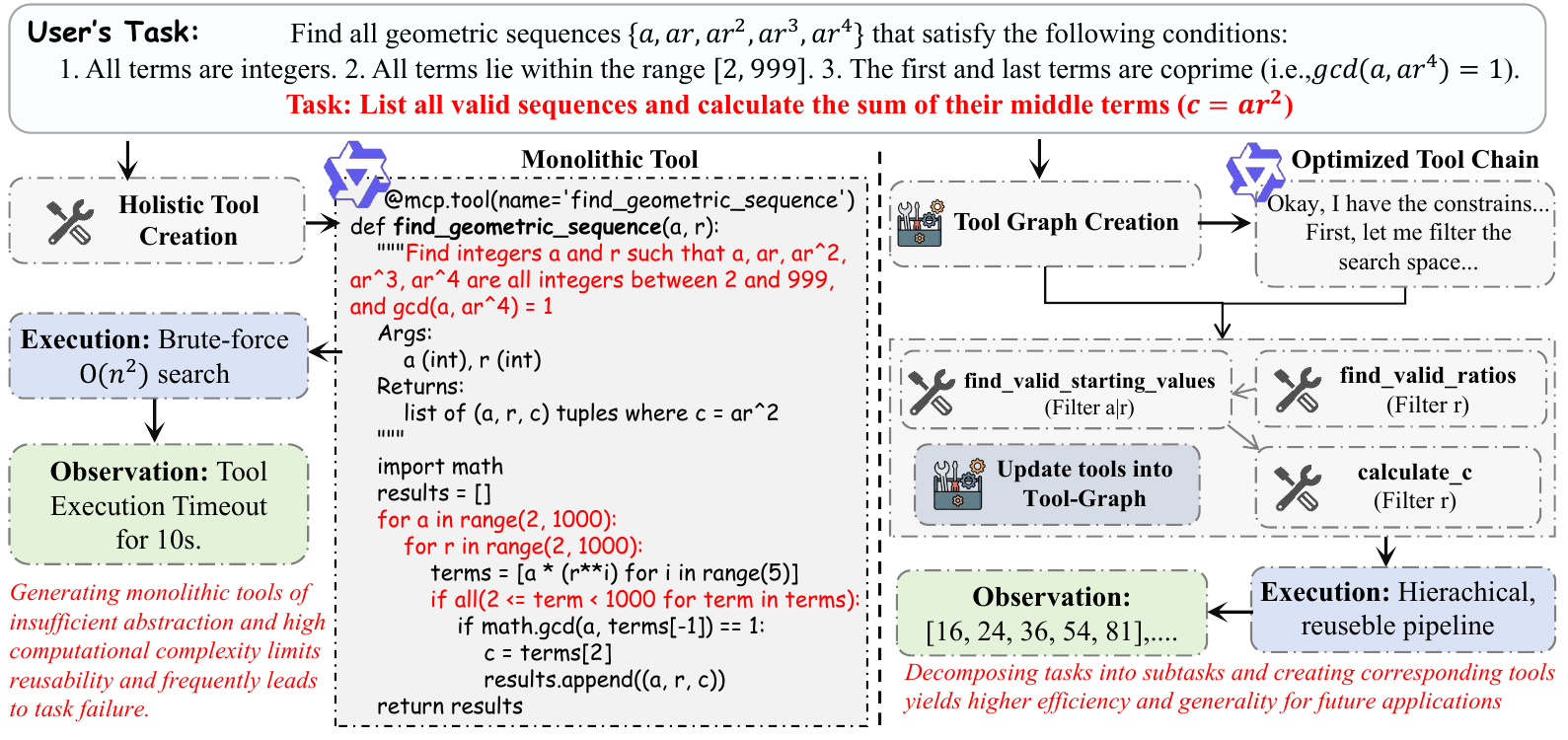}
  \caption{Comparison on constrained geometric sequence optimization. \textbf{(Left)} The baseline uses a monolithic tool causing $O(n^2)$ brute-force timeouts. \textbf{(Right)} Our method employs a modular tool chain to prune the search space, significantly improving efficiency and reusability.}
  \label{fig:case_study}
\end{figure*}

Figure~\ref{fig:case_study} illustrates the paradigm shift enabled by our framework in handling complex reasoning tasks.
In contrast to the baseline, which relies on generating disposable, monolithic tools with high computational complexity ($O(n^2)$), our method produces consistently more accurate and mathematically rigorous solutions through modular tool evolution.
Instead of naively iterating through all possible combinations, our agent identifies key constraints to optimize its reasoning strategy.
It dynamically adjusts its approach by breaking the task into a dependency graph of sub-problems, creating specialized tools for each logical stage.
This divide-and-conquer strategy enables the agent to enforce constraints hierarchically, filtering invalid candidates at the earliest possible step.
As a result, the agent avoids the pitfalls of unstructured brute-force execution, achieving high-efficiency problem solving while populating the tool memory with robust, reusable modules for future tasks.

\subsection{Evolving Dynamics of Tool Graph}
\begin{figure}[!ht]
  \centering
  \includegraphics[width=\linewidth]{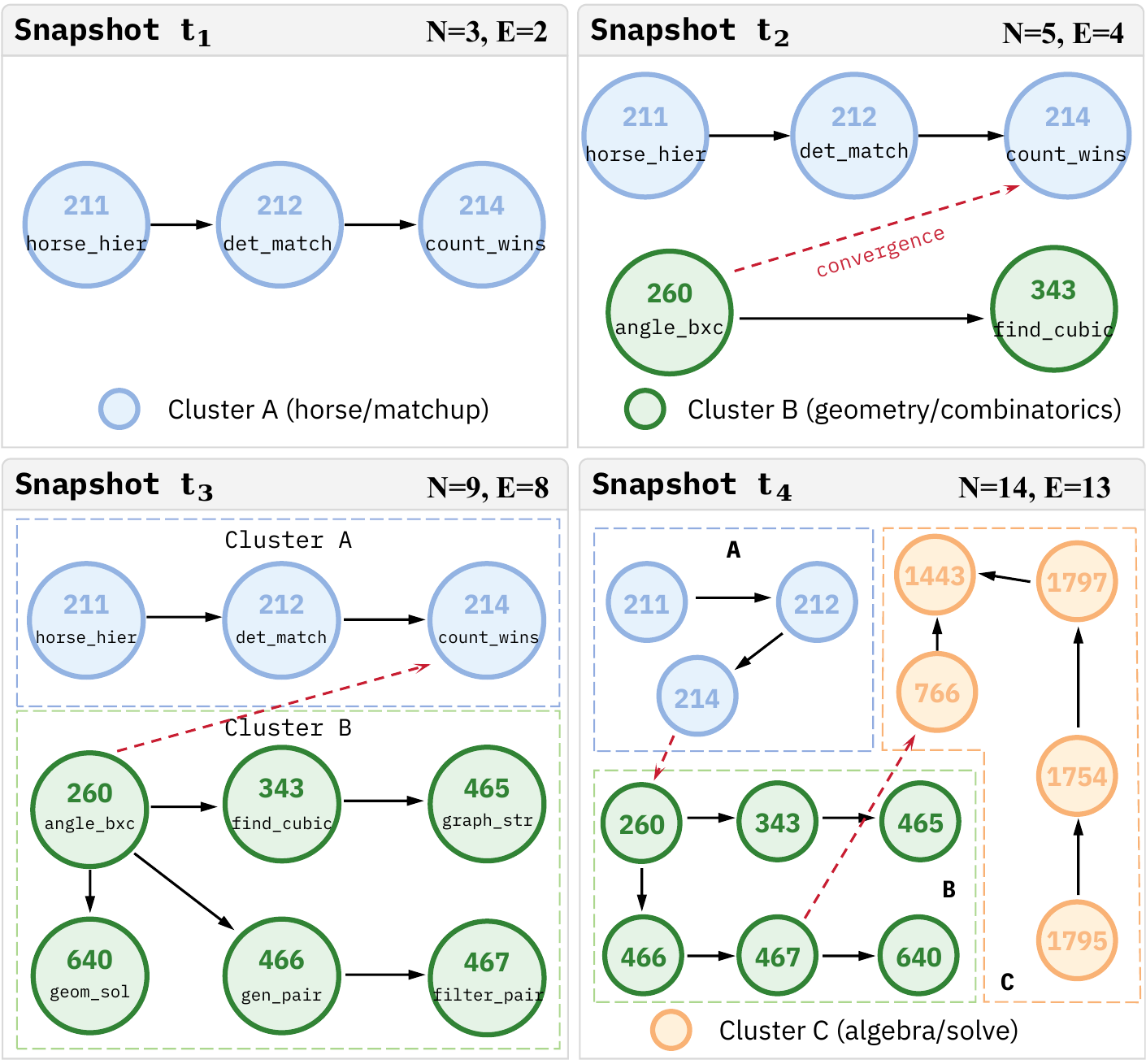}
  \caption{Structural Growth of Tool Subgraphs from Single-Path Reuse to Multi-Branch Convergence.}
  \label{fig:case_study_evolving}
\end{figure}
To better understand the evolving dynamics of our designed tool memory, we extract a representative tool subgraph across four different training steps and analyze its structural progression, as illustrated in Figure~\ref{fig:case_study_evolving}.
Here, $N$ and $E$ denote the number of nodes and edges, respectively, and the black dashed lines indicate the dependencies between tools.
Overall, three distinct functional tool clusters emerge over time. 
During the early stages of training, the tool graph consists of small, disjoint subgraphs. 
As training progresses, these isolated components become connected through tool reuse and merging techniques.
This evolution not only increases the structural complexity of the graph but also integrates experience from different domains. 
Beyond inter-cluster connections, the overall graph complexity also grows significantly, as evident in Snapshot $t_3$. 
In the later stages of training, the graph complexity continues to increase as distinct clusters merge together, ultimately equipping the agent with diverse, cross-disciplinary experiences.

\section{Related Work} \paragraph{Self-Evolving Agents.} Self-evolving agents aim to overcome the static limitations of LLMs through continual adaptation~\cite{gao2025survey}.
Existing research explores adaptation across multiple dimensions, including model parameter updates~\cite{zhou2025self, hu2025agentgen}, long-term memory expansion~\cite{liang2024self, zhang2023large}, and autonomous tool creation or reuse~\cite{qiu2025alita, wang2024toolgen}.
Such evolution occurs either dynamically during inference or via continual learning~\cite{qu2024exploration}.
Notable systems like Alita~\cite{qiu2025alita}, SE-Agent~\cite{lin2025se}, and Agent KB~\cite{agentkb} have demonstrated capabilities in dynamic tool generation, trajectory refinement, and cross-domain knowledge transfer.
Most approaches treat tool evolution and policy learning as independent modules.

\paragraph{Agentic Reinforcement Learning.} Reinforcement learning has become a central paradigm for improving agent decision-making in multi-turn environments, addressing challenges like long-horizon credit assignment and sparse rewards~\cite{wu2025gap, mialon2023gaia, wei2025webagent, zhuang2025workforceagent,xu2026stable}.
While early trajectory-level methods like GRPO~\cite{guo2025deepseek} struggled with coarse feedback signals, subsequent work improved stability through group-based advantage estimation~\cite{feng2025group} and structured rewards that evaluate reasoning quality and tool efficiency~\cite{dong2025tool}.
Recent efforts have scaled training to longer horizons~\cite{gao2025beyond} and integrated hierarchical planning~\cite{zhang2025agent}.
Despite the advances, most methods focus on optimizing policy parameters while neglecting persistent external memory.
We bridge the gap by coupling policy optimization with the growth of a structured \emph{Tool Graph Memory}, allowing agents to refine the decision policy and accumulate durable capabilities simultaneously.

\section{Conclusion} 
We have presented a robust paradigm for building self-evolving agents capable of autonomous tool creation and fine-grained credit assignment. 
Specifically, we introduce the Tool Graph Memory, a dynamic mechanism that not only stores executable tools but also captures their causal dependencies and usage contexts. 
Coupled with anchor-based advantage estimation and designed process rewards, this memory enables the agent to efficiently generalize learned skills to novel tasks. 
Our extensive experiments confirm that this joint optimization of policy and memory yields significant gains.
By enabling agents to build and refine their own Tool Memory over time, this work takes a significant step toward developing truly autonomous, open-ended generalist agents. 

\section*{Limitations}

While \model demonstrates substantial improvements in multi-hop reasoning and remains competitive in mathematical tasks, several limitations persist. 
First, a performance gap remains between our approach and other methods on the GSM8K and MATH500 datasets. 
This suggests that the overhead of generating tools for simple problems may hinder basic reasoning capabilities.
Second, the toolset developed during training may limit the model's adaptability to other contexts, such as direct search or highly specialized domains.
Furthermore, due to the scale of the model, many of the generated tools remain trivial and too simplistic to be effectively reused by other LLMs.
Finally, despite careful design, the reward function may still incentivize superficial reward hacking. 
This underscores the need for further refinement to better align agent incentives with genuine task correctness and reasoning depth.

\section*{Acknowledgements}
We thank the anonymous reviewers and the area chair for their constructive comments. 
The authors of this paper were supported by Shanghai Artificial Intelligence Laboratory.

\bibliography{custom.bib}

\newpage
\appendix
\section{Dataset Statics}
\subsection{Mathematical Reasoning Benchmarks}

\begin{itemize}[leftmargin=1em]
    \item \textbf{AIME24~\cite{aime2024}} serves as a rigorous benchmark for evaluating mathematical reasoning. It comprises 30 challenging problems derived from the American Invitational Mathematics Examination, spanning diverse domains such as algebraic equations and geometric puzzles. Due to its high complexity and the richness of its problem types, AIME24 is widely adopted to assess the reasoning performance of advanced models.

    \item \textbf{MATH500~\cite{lightman2023let}} is a curated subset of 500 challenging problems selected by OpenAI from the larger MATH dataset. These problems cover a broad spectrum of mathematical disciplines, including algebra, geometry, calculus, and number theory, with difficulty levels ranging from high school to collegiate standards. It is frequently used in academic research to evaluate the problem-solving capabilities of various reasoning models.

    \item \textbf{GSM8K~\cite{hendrycks2021measuring}} consists of high-quality grade-school math word problems released by OpenAI. Solving these problems typically requires 2 to 8 steps of multi-step reasoning involving basic arithmetic operations. This dataset is primarily used to test the logical consistency and fundamental mathematical competencies of models.
\end{itemize}
\subsection{Knowledge-Intensive reasoning benchmarks}
\begin{itemize}[leftmargin=1em]
    \item \textbf{HotPotQA~\cite{yang2018hotpotqa}} is a pivotal benchmark for multi-hop question answering. Sourced entirely from Wikipedia, it provides a rich, structured knowledge base designed to evaluate the ability of LLMs to perform complex reasoning and process information across multiple supporting documents.

    \item \textbf{2WikiMultihopQA~\cite{ho2020constructing}} is specifically constructed to assess multi-step reasoning capabilities. It challenges natural language processing models to answer complex queries by integrating and synthesizing evidence from disjoint Wikipedia articles, ensuring that models cannot rely on single-document retrieval alone.

    \item \textbf{MuSiQue~\cite{trivedi2022musique}} serves as a highly challenging benchmark aimed at pushing the boundaries of multi-hop reasoning. By minimizing reasoning shortcuts, it encourages the development of models that go beyond simple information retrieval, requiring deeper semantic understanding and rigorous logical synthesis to derive correct answers.

    \item \textbf{Bamboogle~\cite{press2022measuring}} evaluates reasoning capabilities using ``Google-proof'' questions that resist direct search engine lookup. It focuses on queries where the answer must be derived by combining information from multiple distinct sources. This benchmark is crucial for distinguishing between genuine multi-source synthesis and reliance on parametric memory or simple retrieval heuristics.
\end{itemize}

\section{Baseline Descriptions}
\begin{itemize}[leftmargin=1em]
    \item \textbf{GRPO~\cite{guo2025deepseek}} is a reinforcement learning algorithm based on policy optimization, designed to balance stability, sample efficiency, and theoretical guarantees. By introducing the concept of group-based relative advantage, it simplifies gradient estimation while preserving the theoretical assurance of monotonic policy improvement. GRPO is versatile and applicable to tasks in both continuous and discrete action spaces.

    \item \textbf{DAPO~\cite{yu2025dapo}}, developed by ByteDance Labs, is an RL algorithm tailored to address the stability challenges of large-scale LLM training. It demonstrates superior performance in complex tasks such as mathematical reasoning and code generation. Its proposed ``Clip-Higher'' strategy effectively boosts entropy to encourage sample diversity. Furthermore, DAPO stabilizes the training process through mechanisms like dynamic sampling, token-level policy gradient loss, and overlong reward shaping.

    \item \textbf{REINFORCE++~\cite{hu2025reinforce++}} represents a robust evolution of the classic REINFORCE algorithm, integrating multiple optimization strategies to mitigate high variance. It incorporates baseline subtraction and temporal difference (TD) estimation to stabilize gradient updates, enabling incremental learning without the need to await full trajectories. Additionally, it employs entropy regularization to prevent premature policy rigidity and encourage exploration.

    \item \textbf{ARPO~\cite{dong2025agentic}} is an RL method specifically designed for multi-turn LLM agents. It features an entropy-based adaptive rollout scheme that dynamically intensifies sampling during steps with high uncertainty. Moreover, it incorporates a specialized advantage attribution mechanism to effectively assign credit across complex, branching tool-use interactions.
\end{itemize}
\section{Implementation Details}
\label{appendix:training_details}
In this section, we detail the experimental training settings. 
We implement \model based on the RL-Factory framework~\cite{chai2025rlfactoryplugandplayreinforcementlearning}. 
Crucially, to prevent the model from overfitting to deterministic tool outputs, we exclude tool execution results from the loss calculation; optimization is restricted solely to tokens involved in reasoning and tool invocations. 
For the environment, we utilize the Python sandbox from~\cite{bytedanceseedfoundationcodeteam2025fullstackbenchevaluatingllms} as the coding interface and a local Wikipedia search server~\cite{jin2025search} for retrieval. 
To strike a balance between efficiency and performance, we limit search returns to the top-3 results and impose a 10-second timeout on tool calls to ensure training efficiency.
All experiments are conducted using the Qwen3-4B model in standard generation mode (non-thinking). 
All experiments were performed on NVIDIA H200 GPUs, with each training epoch requiring approximately 10 hours.
Specific hyperparameters are listed in Table~\ref{tab:hyperparameters}.
\begin{table}[h]
    \centering
    \small
    \caption{Agentic RL Training Hyperparameters.}
    \label{tab:hyperparameters}
    \begin{tabular}{l|c}
        \toprule
        \textbf{Hyperparameter} & \textbf{Value} \\
        \midrule
        Backbone Model & Qwen-3-4B \\
        top\_p & 0.98 \\
        rollout\_num & 8 \\
        temperature & 0.7 \\
        repetition\_penalty & 1.05 \\
        max\_turns & 6 \\
        max\_prompt\_length & 4096 \\
        max\_response\_length & 2048 \\
        Global Batch Size & 384 (64 $\times$ 6 GPUs) \\
        Learning Rate & $1.0 \times 10^{-6}$ \\
        Num Train Epochs & 1.0 \\
        \bottomrule
    \end{tabular}
\end{table}

\section{Training Algorithms}
\label{appendix:training_Algorithms}

\begin{algorithm}[H]
\small
\caption{SEARL Training Process}\label{alg:searl}
\begin{algorithmic}[1]
\Require Dataset $\mathcal{D}$, Initial Policy $\pi_{\theta}$, Reference Policy $\pi_{ref}$, Initial Tool Graph $\mathcal{T}_G$, Learning rate $\eta$, Hyperparameter $\lambda$, Iterations $K$, Rollouts per task $N$

\For{iteration $k \gets 1$ \textbf{to} $K$}
    \State Initialize trajectory buffer $\mathcal{B} \gets \emptyset$
    \State Initialize candidate tool buffer $\mathcal{C}_{new} \gets \emptyset$ \Comment{Buffer for newly created tools}
    \State Sample a batch of tasks $X$ from $\mathcal{D}$

    \For{each task $x \in X$}
        
        \For{rollout $i \gets 1$ \textbf{to} $N$}
            \State Initialize trajectory $\tau_i \gets \emptyset$
            
            \For{step $t \gets 1$ \textbf{to} $T$}
                \If{$t == 1$}
                    \State Generate subtask plans $P = [p_1, \dots, p_n]$ based on query $x$
                \EndIf
                
                \State Retrieve tools $T_t \gets \text{Retrieve}(P, \mathcal{T}_G)$
                \State Generate action $a_t \sim \pi_{\theta}(\cdot|s_t, T_t)$
                \State Execute $a_t$, observe reward $r_t$ and next state $s_{t+1}$
                \State Append $(s_t, a_t, r_t, T_t)$ to $\tau_i$
                
                \If{$a_t$ creates a new tool $t_{new}$}
                     \State $\mathcal{C}_{new} \gets \mathcal{C}_{new} \cup \{ t_{new} \}$ \Comment{Collect distinct tools separately}
                \EndIf
            \EndFor
            \State $\mathcal{B} \gets \mathcal{B} \cup \{ \tau_i \}$ \Comment{Collect trajectory for RL}
        \EndFor
        
    \EndFor

    \State \textbf{Memory Evolution:}
    \State Filter valid tools from $\mathcal{C}_{new}$ based on execution success and rewards
    \State Register filtered tools into $\mathcal{T}_G$ via \text{Merge} and consolidation

    \State \textbf{Advantage Estimation:}
    \For{each episode group $\mathcal{G}^E$ in $\mathcal{B}$}
        \State Compute total return $R(\tau) \gets R_{orm} + \sum r_t$
        \State Compute Episode Relative Advantage $A^E$
        
        \State Identify unique MCP tools $\mathcal{U}$ involved in $\mathcal{G}^E$
        \For{each tool anchor $g \in \mathcal{U}$}
            \State Aggregate step-level group $\mathcal{G}_S(g)$ from $\mathcal{B}$
            \State Compute Tool-Anchored Step Advantage $A^S$
        \EndFor
        
        \State Compute final advantage $A_{total} \gets A^E + \lambda \cdot A^S$
    \EndFor

    \State \textbf{Policy Update:}
    \State Optimize $\pi_{\theta}$ by maximizing $\mathbb{E}_{\tau \sim \mathcal{B}}[A_{total} \log \pi_{\theta}]$

\EndFor
\end{algorithmic}
\end{algorithm}

\section{Policy Optimization}
\label{appendix:policy_opt}
Inspired by GiGPO~\cite{feng2025group}, we integrate the two levels of advantage signals into a unified metric for hierarchical credit assignment:
\begin{equation}
    A(a_t^{(i)}) = A^E(\tau_i) + \omega \cdot A^S(a_t^{(i)}),
\end{equation}
where $\omega \in \mathbb{R}_{\ge 0}$ is a weighting coefficient that balances the episode-level and step-level advantages.
Specifically, $A^E(\tau_i)$ evaluates the relative quality of the entire episode compared to others within its group, whereas $A^S(a_t^{(i)})$ provides fine-grained credit assignment for actions taken under analogous tool-use conditions.
Together, they offer robust hierarchical supervision for the policy optimization of LLM agents. Consequently, the clipped policy optimization objective for \model is defined as:

\begin{small}
\begin{equation*}
\begin{split}
&\mathcal{J}_{\text{SEARL}}(\theta)\\ 
&= \mathbb{E}_{\substack{x \sim p(X) \\ \{\tau_i\}_{i=1}^N \sim \pi_{\theta_{\text{old}}}}}
\Biggl[ \frac{1}{NT} \sum_{i=1}^{N} \sum_{t=1}^{T} \min\Bigl( \rho_\theta(a_t^{(i)}) A(a_t^{(i)}),\\
&\mathrm{clip}\bigl(\rho_\theta(a_t^{(i)}), 1 \pm \epsilon\bigr) A(a_t^{(i)}) \Bigr) \Biggr] \\
&- \beta \mathbb{D}_{\mathrm{KL}}\!\bigl(\pi_\theta(\cdot \mid x) \,\|\, \pi_{\mathrm{ref}}(\cdot \mid x)\bigr),
\end{split}
\end{equation*}
\end{small}
where $\rho_\theta(a_t^{(i)}) = \frac{\pi_\theta(a_t^{(i)} \mid s_t^{(i)}, x)}{\pi_{\theta_{\text{old}}}(a_t^{(i)} \mid s_t^{(i)}, x)}$ is the importance weight, and $\beta$ scales the KL penalty that enforces proximity to the reference policy $\pi_{\text{ref}}$.

\section{Tool Creation and Retrieval Details}
\label{appendix:training_Retrieval}

\begin{lstlisting}[language=Python,
caption={MCP Creation and Execution Tool},
label={code:mcp_tool},
showspaces=false,
showstringspaces=false
]
@mcp.tool()
def create_and_execute_mcp(
    name: str,
    description: str,
    arguments: str,
    returns: str,
    code: str,
    inputs: Dict[str, Any],
    timeout: float = 15.0
) -> str:
    """
    MCP Creation and Execution Tool

    Create and immediately execute an MCP tool function.

    Args:
        name: MCP tool name (Function name)
        description: Tool description
        arguments: Argument description string (e.g., "a, b (int)")
        returns: Return value description
        code: Complete Python function implementation code
        inputs: Input arguments dictionary required for this function call (e.g., {"a": 1, "b": 2})
        timeout: Execution timeout in seconds (default: 15.0)

    Returns:
        str: JSON formatted string containing creation status and execution result
             {
                 "creation_success": bool,
                 "execution_result": any,
                 "stdout": str,
                 "stderr": str,
                 "error": str (optional)
             }
    """
    ...
\end{lstlisting}
\subsection{Tool Creation}
We enable the agent with the power of tool creation by introducing a predefined tool named~\texttt{mcp creation tool}, thus utilizing the basic tool-calling ability of LLMs.
The detailed code input can be found in Code~\ref{code:mcp_tool}.
Different from solely creating tools, we directly execute the created tool to save reasoning steps in LLMs, and filter out the creation failed tools.

Upon creation, the generated tools $\texttt{t}_i$ are not immediately registered in the global \texttt{Tool Graph} $\mathcal{T}_G$ to mitigate the risk of duplication across different rollouts of the same training sample.
Instead, during the formal registration phase, we aggregate all trajectories $[\tau_1, \tau_2, \dots, \tau_N]$ associated with the sample.
We compute the cumulative reward for each individual trajectory---summing both outcome and process scores---and retain only the optimal trajectory $\tau_i$ with the highest reward.
Subsequently, the extracted sub-plan graph $G_{\text{plan}}$ and its corresponding tools are integrated into $\mathcal{T}_G$.
In this structure, tools represent nodes while subtask dependencies serve as edges, forming a \textbf{connected component}.
Notably, we employ an embedding model to encode the description of tool $v$ into a semantic embedding $\textbf{e}(v)$, which is then integrated as a feature of the node.

\subsection{Tool Retrieval}
In contrast to the complex lifecycle of tool creation, spanning invocation, generation, and registration, the tool retrieval procedure is significantly more streamlined.
This process occurs exclusively during the system prompt generation stage, following the computation of plans.
Given the derived subtask plan list $[p_1, p_2, \dots, p_n]$, we first encode the textual description of each plan into embeddings $[\textbf{e}_p^1, \textbf{e}_p^2, \dots, \textbf{e}_p^n]$.
We then utilize a retrieval model $\mathcal{R}$ to identify the top-$k$ relevant tools from the Tool Graph $\mathcal{T}_G$ via graph traversal.
Finally, these tools are appended to the system prompt.
Crucially, this mechanism serves as a recommendation rather than a constraint; the agent retains the autonomy to decide whether to invoke the retrieved tools.
The detailed prompt for tool recommendation can be found in Appendix~\ref{appendix:prompt}.

\subsection{Retrieval Model}
To ensure robust and contextually relevant experience retrieval, we implement a \textbf{Hybrid Retrieval} framework that synthesizes the strengths of both sparse and dense retrieval mechanisms.
This dual-stream approach captures relevance at different levels of abstraction:

\paragraph{Sparse Retrieval (Text-based).}
For surface-level term matching, we utilize traditional information retrieval techniques based on TF-IDF (Term Frequency-Inverse Document Frequency).
This method represents textual content as sparse, high-dimensional vectors, quantifying the importance of terms relative to the corpus.
It excels at identifying documents with significant keyword overlap, ensuring high precision when vocabulary alignment is strong.

\paragraph{Dense Retrieval (Semantic).}
To capture deeper contextual relationships beyond exact keyword matching, we employ a dense retrieval component.
Specifically, we utilize the \texttt{sentence-transformers/all-MiniLM-L6-v2} model, a lightweight transformer-based encoder that maps sentences into a continuous vector space.
By computing cosine similarity between embeddings, this method retrieves experiences that are semantically related even in the absence of lexical overlap.

\paragraph{Hybrid Fusion.}
To mitigate the limitations of individual methods, we fuse the results using a weighted ranking strategy.
For a retrieved experience $e_i \in \mathcal{E}$, the final relevance score $\sigma_i^{\text{hyb}}$ is computed as a linear combination of the sparse score $\sigma_i^{\text{text}}$ and the dense score $\sigma_i^{\text{sem}}$:
\begin{equation}
    \sigma_i^{\text{hyb}} = \alpha \cdot \sigma_i^{\text{text}} + (1 - \alpha) \cdot \sigma_i^{\text{sem}},
\end{equation}
where $\alpha \in [0, 1]$ is a tunable parameter (setting $\alpha = 0.5$ in our settings) that balances the trade-off between lexical precision and semantic generalization.
This hybrid mechanism ensures robustness against both syntactic variation and conceptual drift.

\section{Prompt}
\label{appendix:prompt}
\begin{tcolorbox}[
    enhanced,
    float*=t,
    width=\textwidth,
    colback=white,             
    colframe=black,            
    boxrule=0.8pt,             
    arc=3pt,                   
    colbacktitle=black,        
    coltitle=white,            
    fonttitle=\bfseries,       
    title=Initial Plan Generation, 
    sharp corners=south,       
    left=2mm,
    right=2mm,
    top=2mm,
    bottom=2mm
]
\small 
\texttt{Create a step-by-step plan for the given task.}\\
\textbf{Instructions:}
\begin{enumerate}
    \item \texttt{Break the task into subtasks (ST1, ST2, ST3, ...). If the task is indivisible, use only ST1.}
    \item \texttt{Define subtask dependencies in a \textbf{\#\#DAG\_LIST}. Example: \texttt{[(ST1, ST2)]} means ST2 depends on ST1. For a single task, use \texttt{[(ST1)]}.}
    \item \texttt{For each subtask \textbf{\#\#STn}, provide clear, actionable steps.}
    \item \texttt{The plan must only contain concrete actions. Do not include explanations or simulated code.}
\end{enumerate}

\texttt{SUBTASKS EXAMPLE:}
\begin{verbatim}
##DAG_LIST
[(ST1, ST3), (ST1, ST2), (ST2, ST3)]
##ST1:xxx
1. xxx.  
2. xxx.
##ST2:xxx
1. xxx.  
2. xxx.  
3. xxx.
4. xxx.
##ST3:xxx
1. xxx.  
2. xxx.  
3. xxx.
\end{verbatim}

\texttt{Previous is an example of generating subtasks, you are not required to solve the task within the plan itself; focus on outlining the steps.}\\
\texttt{Always verify your answers before providing them, provide the plan for verifying.}\\
\texttt{If you are uncertain about the task, you can plan for web search to gather more information.}\\

\vspace{0.1cm}
\texttt{Now, write a plan below to solve the task within \{\{max\_turns-1\}\} steps.}
\end{tcolorbox}

\begin{tcolorbox}[
    enhanced,
    float*=t,
    width=\textwidth,
    colback=white,             
    colframe=black,            
    boxrule=0.8pt,             
    arc=3pt,                   
    colbacktitle=black,        
    coltitle=white,            
    fonttitle=\bfseries,       
    title=System Prompt: Tool Creation Prompt, 
    sharp corners=south,       
    left=2mm,
    right=2mm,
    top=2mm,
    bottom=2mm
]

\lstset{
    basicstyle=\small\ttfamily, 
    breaklines=true,            
    breakatwhitespace=false,    
    columns=fullflexible,       
    keepspaces=true,            
    showstringspaces=false,     
    frame=none,                 
    aboveskip=0pt,
    belowskip=0pt
}

\begin{lstlisting}
You are a step-by-step problem solver. For each step, follow this loop:
- <subtask> ST_X: {step} </subtask>
- <thinking> ... </thinking>: explain reasoning, note whether an MCP tool is needed
- <tool_call> ... </tool_call>: use plain Python if no MCP is needed; to create a new MCP, call create_and_execute_mcp tool.
Examples:
- Create a reusable MCP:
  <subtask> ST_1: Build quadratic solver MCP </subtask>
  <thinking> I need to create a quadratic equation solver to handle equations of the form ax^2+bx+c=0. </thinking>
  <tool_call>
  {
      "name": "create_and_execute_mcp",
      "arguments": {
          "name": "quadratic_solver",
          "description": "Solve ax^2+bx+c=0",
          "arguments": "a,b,c (float)",
          "returns": "roots list",
          "code": "def quadratic_solver(a,b,c):\n    import math\n    d=b*b-4*a*c\n    if d>0:\n        r1=(-b+math.sqrt(d))/(2*a)\n        r2=(-b-math.sqrt(d))/(2*a)\n        return [r1,r2]\n    if d==0:\n        r=-b/(2*a)\n        return [r,r]\n    return 'complex roots'",
          "inputs": {"a": 1, "b": -3, "c": 2}
      }
  }
  </tool_call> 
Rules:
- Only create an MCP when it is reusable; at most one MCP per step; ensure a unique MCP name
- Do not reuse tool names as variables; each code block is state-isolated-redefine imports/vars/functions every step
- Repeat the cycle of <subtask> ST_X: {step} </subtask> <thinking> ... </thinking> <tool_call> ... </tool_call> until the task is solved.
- Final answer must be in <answer>\boxed{...}</answer> with only the boxed result
--- Here is the Task and Plan: ---
Task:
**{{question}}**
Plan:
**{{input_plan}}**
Now begin to solve the task according to the given plan. Now begin with ST_1. 
If you solve the task correctly, you will receive a reward of $1,000,000. 

\end{lstlisting}

\end{tcolorbox}

\begin{tcolorbox}[
    enhanced,
    float*=t,
    width=\textwidth,
    colback=white,             
    colframe=black,            
    boxrule=0.8pt,             
    arc=3pt,                   
    colbacktitle=black,        
    coltitle=white,            
    fonttitle=\bfseries,       
    title=System Prompt: Tool \& Reasoning Format, 
    sharp corners=south,       
    left=2mm,
    right=2mm,
    top=2mm,
    bottom=2mm
]

\lstset{
    basicstyle=\small\ttfamily,
    breaklines=true,
    columns=fullflexible,
    keepspaces=true,
    frame=none
}

\begin{lstlisting}
You are a helpful assistant who can solve the given question step by step with the help of available tools. Given a question, you need to think about the reasoning process and then provide the answer. While thinking, you can invoke tools to search for information or perform calculations as needed.
# Reasoning and Answer Format
- The reasoning process should be enclosed within <think> </think> tags (if thinking mode is enabled).
- The final answer must be enclosed within <answer> </answer> tags.
- The final exact answer should be enclosed within \boxed{} with LaTeX format inside the <answer> tags.
# Example
<think>
I need to search for information about this topic first.
</think>
<tool_call>
{
    "name": "search",
    "arguments": {
        "query": "search query here"
    }
}
</tool_call>
[Tool execution result will be automatically returned here]
<think>
Now I have the information, let me calculate the result using Python.
</think>
<tool_call>
{
    "name": "execute_python_code",
    "arguments": {
        "code": "print(2+3)",
        "timeout": 10.0
    }
}
</tool_call>
[Tool execution result will be automatically returned here]
<think>
Based on the search results and calculations, I can now provide the final answer.
</think>
<answer>
The final answer is \boxed{answer here}
</answer>
Note: Tool execution results are automatically returned by the system. You do not need to include <tool_response> tags. Simply continue with your reasoning after the tool call.
\end{lstlisting}

\end{tcolorbox}

\end{document}